\documentclass[a4paper,conference]{IEEEtran}
\usepackage{cite}
\usepackage[misc]{ifsym} 

\ifCLASSINFOpdf
   \usepackage[pdftex]{graphicx}
   \graphicspath{{../pdf/}{../jpeg/}}
  \DeclareGraphicsExtensions{.pdf,.jpeg,.png}
\else
\fi
\usepackage{amsmath}
\usepackage{amsfonts}
\usepackage{bbm}
\usepackage{array}
\usepackage{booktabs}
\usepackage{relsize}
\usepackage{multirow}

\begin{document}
%
\title{MOST-Net: A Memory Oriented Style Transfer Network for Face Sketch Synthesis}

\author{\IEEEauthorblockN{Fan Ji\textsuperscript{1}, Muyi Sun\textsuperscript{1}, Xingqun Qi\textsuperscript{2}, Qi Li\textsuperscript{1}, Zhenan Sun\textsuperscript{1, \textrm{\Letter}}}
\IEEEauthorblockA{\textsuperscript{1}CRIPAC \& NLPR, 
Institute of Automation,
Chinese Academy of Sciences, Beijing, China, 100190\\
\textsuperscript{2}School of AI/Auto, Beijing University of Posts and Telecommunications, Beijing, China, 100876\\
Email: \{fan.ji, muyi.sun\}@cripac.ia.ac.cn,  
xingqunqi@bupt.edu.cn,
\{qli, znsun\}@nlpr.ia.ac.cn}}


%


\maketitle

\begin{abstract}
Face sketch synthesis has been widely used in multi-media entertainment and law enforcement.
Despite the recent developments in deep neural networks, accurate and realistic face sketch synthesis is still a challenging task due to the diversity and complexity of human faces.
Current image-to-image translation-based face sketch synthesis frequently encounters over-fitting problems when it comes to small-scale datasets.
To tackle this problem, we present an end-to-end Memory Oriented Style Transfer Network (MOST-Net) for face sketch synthesis which can produce high-fidelity sketches with limited data. 
Specifically, an external self-supervised dynamic memory module is introduced to capture the domain alignment knowledge in the long term.
In this way, our proposed model could
obtain the domain-transfer ability by establishing the durable relationship between faces and corresponding sketches on the feature level.
Furthermore, we design a novel Memory Refinement Loss (MR Loss) for feature alignment in the memory module, which enhances the accuracy of memory slots in an unsupervised manner.
Extensive experiments on the CUFS and the CUFSF datasets show that our MOST-Net achieves state-of-the-art performance, especially in terms of the Structural Similarity Index(SSIM).

\end{abstract}


%
\IEEEpeerreviewmaketitle

\section{Introduction}
Face sketch synthesis refers to generating sketches from face photos, which attracts extensive attention in recent years. 
Face sketch synthesis plays an important role in a wide range of applications including digital entertainment, law enforcement, and face de-identification\cite{wang2014comprehensive}. 
For instance, people tend to use sketches instead of real photos as an avatar on social media to protect their privacy. 
Nevertheless, it will take a vast time and effort of a professional artist to draw a dedicated sketch. 
The automatic sketch synthesis techniques are highly necessary. 
Early face sketch synthesis methods can be roughly divided into two categories: exemplar-based and regression-based methods. 
Exemplar-based methods \cite{tang2003face,chang2010face,zhang2019deep} 
generate sketches by mining the relationship between corresponding input image patches and sketch patches of face-sketch pairs in the training set. 
However, the synthesized sketches are limited to holding the fidelity and identifiability of the face photo.
Regression-based approaches formulate a linear mapping between photos and sketches as a regression problem\cite{zhou2012markov,meer1991robust}. 
However, the mapping function is not comprehensive and accurate because of the limited capability of the linear regression strategies\cite{liu2005nonlinear}.

\begin{figure}[!t]
\centering
\includegraphics[scale=0.4]{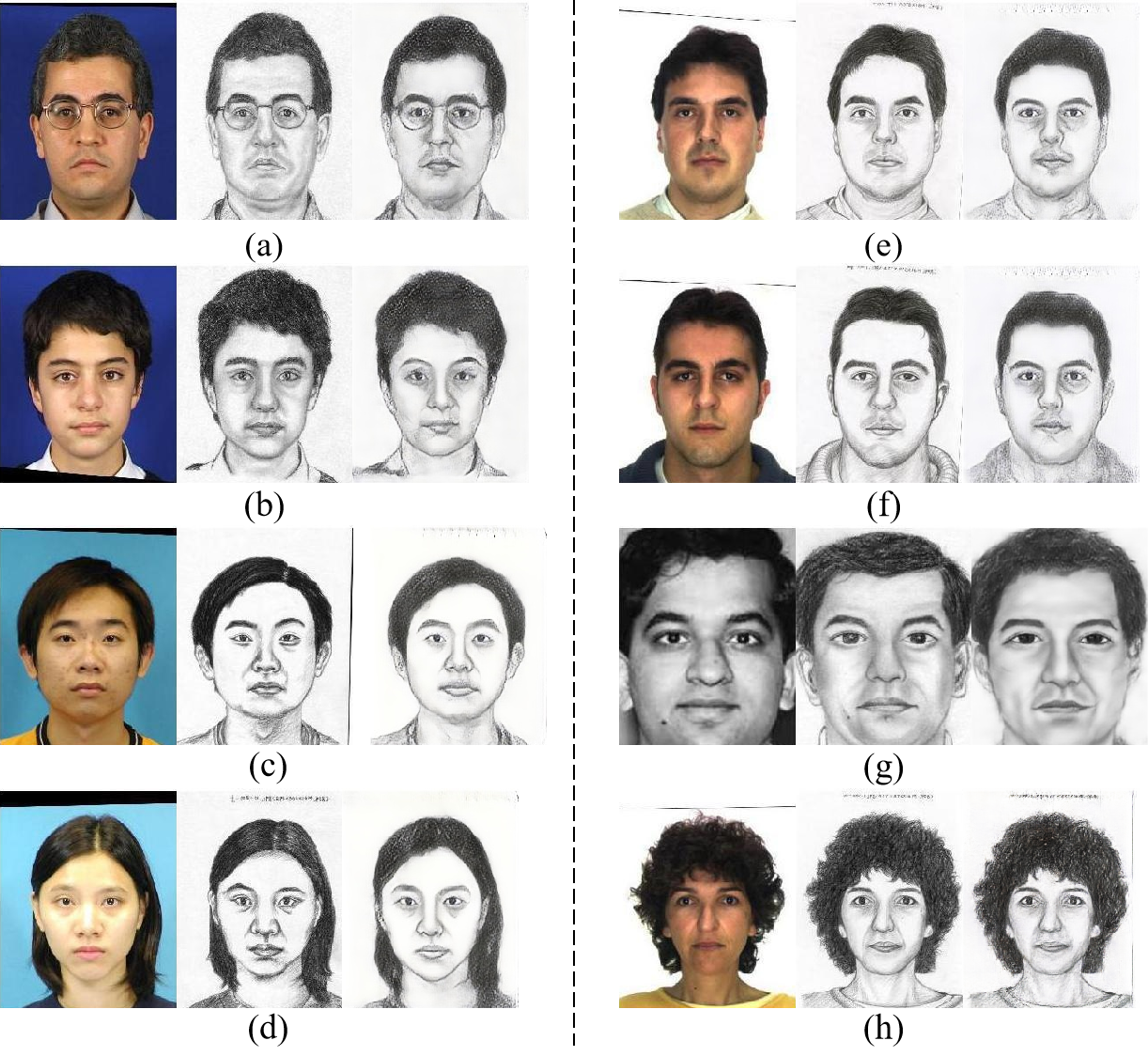}
\caption{Samples of face-sketch pairs and sketches synthesized by the proposed MOST-Net on different databases:(a) (b) XM2VTS database, (c) (d) CUHK database, (e) (f) AR database, (g) CUFSF database. For each sample, from the left to right are real face photo, real sketch, and synthesized sketch. Additionally, (h) is generated by Pix2Pix model.}
\label{fig:task}
\end{figure}

The last few years have seen significant growth in the use of Generative Adversarial Networks (GANs)\cite{goodfellow2014generative} for image generation tasks. 
Face sketch synthesis has also undergone substantial and fundamental improvement. 
A great number of researchers employ various GAN models on this task\cite{zhang2015end,isola2017image,yu2020toward}. Although desirable results are obtained, these approaches often encounter the over-fitting problem with the limited data. 
Take the Pix2Pix \cite{isola2017image} model as an example, we could observe that the generator tries to remember all training data instead of learning the transferring strategy. As shown in Fig. \ref{fig:task}, a string of letters appears in the upper part of the generated sketch as well as the real sketch. However, there should be no letters that do not exist in the input face photo.
We infer that this is because real sketches are only used as the L1 loss supervision in the data-limited tasks. 
To make better use of the sketch knowledge to improve the generalization capacity of the network, the real sketch itself or its features could also be employed as a part of the input, not just for supervision. 


To prevent the over-fitting problem, we propose a novel Memory Oriented Style Transfer Network for Face Sketch Synthesis (MOST-Net). 
Firstly, a face encoder and a sketch encoder are utilized to obtain the face feature maps and the sketch feature maps of the same size. 
Secondly, inspired by \cite{yoo2019coloring}, we construct an external memory block to store two different types of information: key memory and value memory.
The key memory K stores information about the face features, which are used to compute the cosine similarity with input queries. 
The value memory V stores information about the sketch features, which are later used as the condition for the sketch synthesis network. 
This memory module can be described as a reference dictionary to find the corresponding sketch features. 
Meanwhile, we proposed a novel Memory Refinement Loss (MR Loss) for memory alignment with unsupervised learning. 
Thirdly, we inject the sketch features as style information into the generation pipeline of MOST-Net. 
As a result, we recombine the sketch feature slots in the memory through the queries of image features for the final sketch synthesis, which could overcome the over-fitting problem caused by singly end-to-end supervised training.
And in the inference time, we only input the face images as queries and employ the memory to provide sketch-related features for the feature-level fusion.
Consequently, our method achieves state-of-the-art performance in a variety of metrics on the CUFS and the CUFSF datasets.

The contributions of this paper can be summarized as:

\begin{itemize}
\item To the best of our knowledge, the proposed MOST-Net is the first face sketch synthesis network augmented by external neural memory networks.
\item A novel Memory Refinement Loss (MR Loss) is proposed for feature alignment in the memory module. The MR Loss ensures that the key value pair of the memory network can be updated correctly according to the spatial information.
\item Extensive experimental results on CUFS and CUFSF datasets show the superiority of our method.
\end{itemize}

\section{Related Work}

\subsection{Deep Learning-Based Face Photo-Sketch Synthesis}
Deep learning-based face photo-sketch synthesis has developed rapidly in recent years.
Zhang \textit{et al.}\cite{zhang2015end} pioneered an end-to-end fully convolutional neural network to generate sketch. 
Limited by the shallow network architecture and pixel-level loss, its performance is not high-fidelity. 
Isola \textit{et al.}\cite{isola2017image} proposed a conditional GAN (cGAN), called Pix2pix,  for general image-to-image translation tasks. 
Coincidentally, another profound method CycleGAN \cite{zhu2017unpaired} enforced image to image translation by introducing a cycle consistency loss. 
Both of these two general backbone networks can be applied to fake sketch synthesis. 
Following the Pix2pix model and CycleGAN, researchers have made extensive progress. 
On the one hand, several works focused on improving the capability of the GAN models. 
Wang \textit{et al.}\cite{wang2018high} proposed multi-scale generator and discriminator architectures for synthesizing sketches. 
However, undesirable artifacts and distorted structures still exist in many scenarios. 
On the other hand, many researchers found that introducing prior information can effectively improve the performance of the sketch synthesis methods. SCAGAN\cite{yu2020toward} introduced face parsing layouts as input of cGAN to achieve state-of-the-art performance. 
Qi \textit{et al.}\cite{qi2021face} decomposed face parsing layouts into multiple compositions as the semantic-level spatial prior for style injection. 
Nevertheless, besides the model becoming more complex, they could not guarantee that the prior information is always available and beneficial.

\subsection{Memory Network}
The external memory module is used to store critical information over long periods. 
As a typical recurrent neural network (RNN), long short-term memory (LSTM)\cite{hochreiter1997long,graves2012long} is the most classical neural network with a memory mechanism, which dominated the field of sequential data processing. 
Weston \textit{et al.}\cite{weston2014memory} described the memory network as a dynamic knowledge base of question answering. 
Thus, knowledge is compressed into dense vectors to improve the neural network performing memorization. Recent research has shown that the memory network can be applied to various tasks, not only in the field of natural language processing. 
Yoo \textit{et al.}\cite{yoo2019coloring} presented a memory-augmented colorization model to produce high-quality colorization with limited data. 
Huang \textit{et al.}\cite{huang2021memory} performed image de-raining through memory-oriented semi-supervised method. 
Inspired by previous research, we propose a self-supervised memory mechanism for face sketch synthesis.

\subsection{Image Style Transfer}
Initial style transfer aims to transform photos into painting-like images, which indicates that face sketch synthesis can be regarded as a style transfer task. 
Gatys \textit{et al.}\cite{gatys2016image} employed the VGG network \cite{simonyan2014very} in separating image content from style, then the semantic content can be re-rendered in different styles. 
Ulyanov \textit{et al.}\cite{ulyanov2016texture,ulyanov2017improved} found that Instance Normalization (IN) is far more effective than Batch Normalization (BN) in the field of style transfer. 
According to his research, the following studies on style transfer tends to be closely related to IN. 
Consecutively, Huang \textit{et al.}\cite{huang2017arbitrary} proposed Adaptive Instance Normalization (AdaIN) for real-time arbitrary image style transfer. 
In the meanwhile, NVIDIA proposed spatially-adaptive normalization (SPADE) \cite{park2019semantic} which injects style information into the given layout to obtain photorealistic images. 
The SPADE has been applied to various image synthesis tasks, including face image editing\cite{zhu2020sean}, face reenactment\cite{hao2020far}, and person image generation\cite{lv2021learning}. 
In this paper, we also borrow ideas from SPADE for sketch synthesis.

\begin{figure*}[htb]
    \centering
    \includegraphics[scale=0.54]{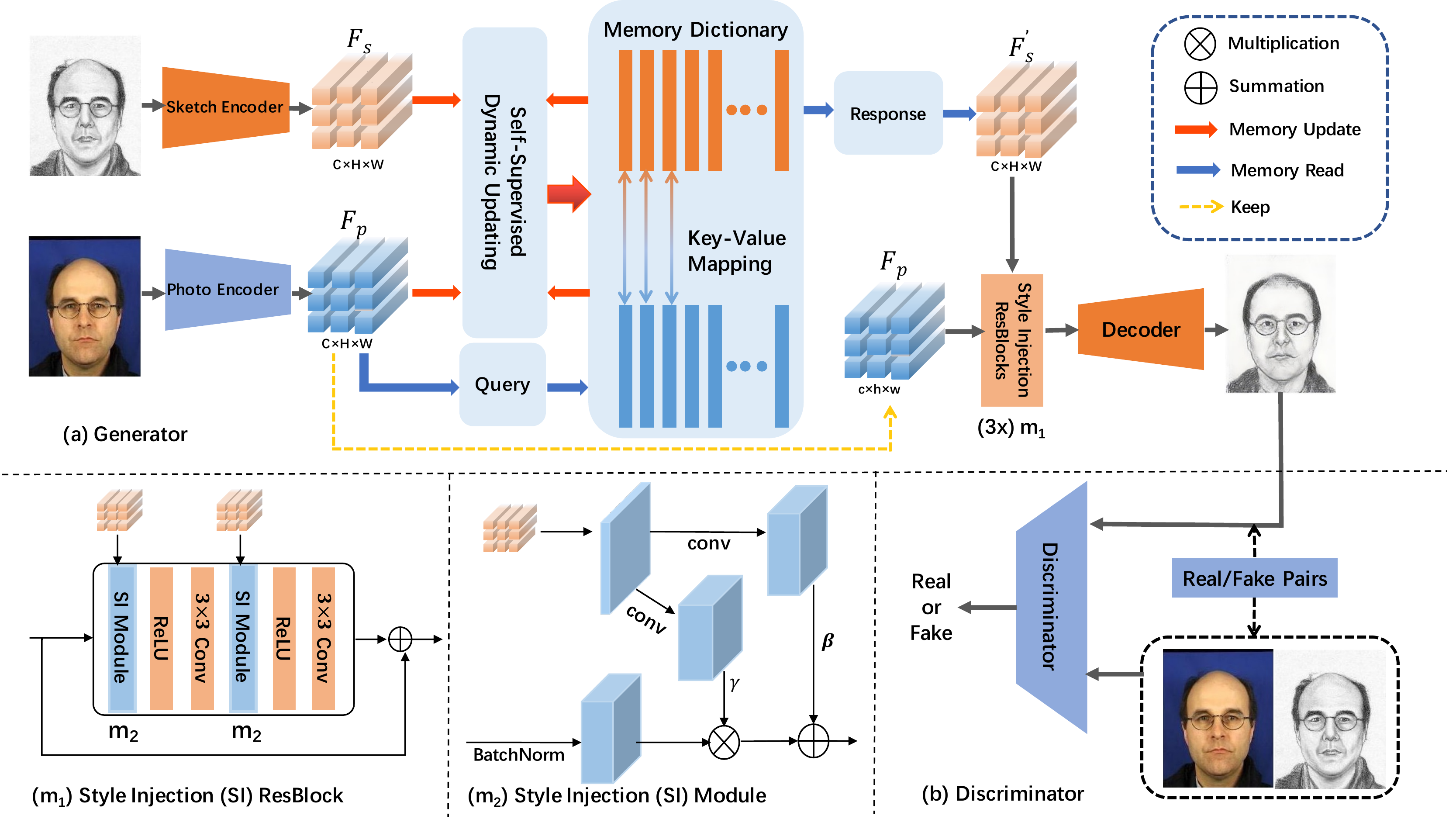}
    \caption{The pipeline of end-to-end Memory Oriented  Style Transfer Network (MOST-Net). (a) The generator architecture of MOST-Net, which contains a sketch encoder, a face photo encoder, a memory dictionary, a decoder and several Style Injection (SI) ResBlocks. (b) The discriminator of MOST-Net. (M1) Details of the Style Injection ResBlock. (M2) Details of Style Injection (SI) Module. In each Style Injection Module, the style feature map is convoluted to produce pixel-level normalization parameters $\gamma$ and $\beta$. }
    \label{fig:Architecture}
\end{figure*}

\section{Method}
Theoretically, SPADE \cite{park2019semantic} is an ideal image domain transfer method. However, in terms of the face sketch synthesis task, there is no such aligned domain-variant image as layout during the inference and test phase. Therefore, our memory-oriented network aims to construct a Memory Dictionary that maps images from photo domain $P$ to sketch domain $S$ on the feature level over a long period of time. Through this memory dictionary, we can still obtain this domain-variant layout information without any input sketch in the testing. 
On the other hand, previous research synthesized sketches directly with the input of only real photos, in which sketches serve as supervision. Our method is able to take both photos and sketches as inputs of our end-to-end network in the training phase, which could take full advantage of ground-truth images. 
Fig.\ref{fig:Architecture} illustrates the overall architecture of the proposed Memory Oriented Style Transfer Network (MOST-Net) for face sketch synthesis. In the following, a detailed description of proposed method is given.

\subsection{Network Architectures}\label{networkArchitectures}
As shown in the Fig.\ref{fig:Architecture}, the proposed MOST-Net consists of a Photo Encoder $PE$, a Sketch Encoder $SE$, a Memory dictionary $M$, three Style Injection (SI) resblocks, and a Decoder $D$. 

Given paired photo-sketch samples shown in Fig.1 $\left\{\left(x_{i},y_{i}\right)\mid x\in \mathbb{R} ^{3\times H\times W}, y\in \mathbb{R} ^{1\times H\times W}\right\}_{i=1}^{N} $,
SE and PE firstly extract photo features $F_{p} \in \mathbb{R}^{c\times h \times w}$ and sketch features $F_{s} \in \mathbb{R}^{c\times h \times w}$, where $x_{i}$ represents the photo, $y_{i}$ represents the sketch, and $N$ represents the total number of samples.
The SE and PE share the same structure which consists of a convolution layer and a stack of residual blocks. 
The memory dictionary is located after the encoders. According to Fig.\ref{fig:Architecture}, there are three main data flows painted with different colors.
The data flow in red indicates that the memory dictionary $M$ is updated in a self-supervised manner when new training samples are fed in. 
The blue one shows that $F_{p}$ serves as a query to retrieve the most relevant slots in $M$.

Detailed updating and retrieving processes are presented in Sec. \ref{memory}. 
Next, we design three SI Resblocks, which inject $\hat{F_{s}}$ as 
domain style information into $F_{p}$ to produce fake sketch representation $z\left(\hat{y}\right)$. 
As shown in Fig. \ref{fig:Architecture} $(M_{1})$ and $(M_{2})$, a SI Resblock takes $F_{p}$ and $\hat{F_{s}}$ as input, which are then processed through stacks of spatial normalization, activation, and convolution operations.
Motivated by \cite{park2019semantic}, our method considers $\hat{F_{s}}$ as style semantic information for realistic sketch synthesis.
Finally, the decoder $D$ predicts fake sketches from $z\left(\hat{y}\right)$, which consists of a stack of residual blocks followed by a convolutional output layer.  
Additionally, a Discriminator is employed in the same setting in Pix2Pix\cite{isola2017image}.


\subsection{Memory Dictionary}
\label{memory}

We define the memory mechanism as a dictionary $M$ holding pairs of the key $k_{i}\in \mathbb{R}^{c}$ and value  $v_{i}\in \mathbb{R}^{c}$, which represents the photo feature and its corresponding sketch feature. Each entry $(k_{i},v_{i})$ of $M$ represents a feature mapping from photo domain to sketch domain. This mapping function indicates the domain transferring strategy on the feature level. We consider the value corresponding to the key closest to each input photo feature to be the expression of this photo feature in the sketch domain. In this way, during model training, we simultaneously input pairs of features to optimize the memory dictionary. In the generation stage, the photo feature is used as the query to obtain the sketch feature. 


Therefore, a dictionary $M$ containing K items can be described in the following:
\begin{equation}\label{1}
M = \left \{ (k_{i},v_{i})\mid k_{i},v_{i}\in\mathbb{R}^{c},i=1,2,3\cdots ,K \right \}
\end{equation}
Additionally, $M$ is initialized randomly, because our method do not rely on any prior knowledge, such as face parsing and salience detection.

According to Sec.\ref{networkArchitectures}, the photo encoder and the sketch encoder convert input photos and sketches to photo features and sketch features. Given photo features $F_{p}$ and sketch features $F_{s}$, we divide them into a number of slots. For example, a face feature slot is the basic unit of complete face feature maps, which serves as a query to dictionary $M$. Specifically, 
\begin{equation}\label{2}
F_{p}= \left \{ f_{1}, f_{2},f_{3},\cdots f_{N}\right \},f_{i}\in \mathbb{R}^{c} 
\end{equation}
\begin{equation}\label{3}
F_{s}= \left \{ s_{1}, s_{2},s_{3},\cdots s_{N}\right \},s_{i}\in \mathbb{R}^{c}
\end{equation}
\begin{equation}
    N =  h\times  w 
\end{equation}
where $f_{i}$ represents a face feature slot, $s_{i}$ represents a sketch feature slot, $N$ is the number of slots (feature map size), $c$ denotes the dimension of the each slot (the number of the feature map channels). Since these slots are intended to interact with Memory Dictionary for reading and updating, they have the same dimension with keys and values.

Our memory $M$ is updated after a photo-sketch pair is introduced to the networks in the training stage. We employ $F_{p}$ to update the keys and $F_{s}$ to update the values in $M$ respectively using the same strategy. For each face feature slot $f_{i}$, we firstly find the most relevant key $k_{max}$ in $M$ through cosine similarity which is illustrated as

\begin{equation}
k_{max}  = \mathop{\arg\max}\limits_{j}(\frac{f_{i}\cdot  k_{j}}{\left \| f_{i} \right \| \left \| k_{j} \right \| })
\end{equation}
Then we update the memory key $k_{max}$ by
\begin{equation}\label{6}
    k_{max}\longleftarrow \alpha\cdot  k_{max} + (1-\alpha )\cdot f_{i}
\end{equation}
where $\alpha \in \left [ 0,1 \right ] $ is a decay rate. Also, each sketch feature slot $s_{i}$ is used to update the values of $M$ with the same strategy as above.  
During each iteration, no additional labels are required. Though we do not assign face feature slots and sketch feature slots to update specific keys and values, a key-value pair $(k_{i},v_{i})$ in $M$ still represents the mapping from the photo feature slot to the corresponding sketch feature slot.
Therefore, the proposed memory dictionary is trained in a self-supervised manner, which is also called as self-supervised dynamic updating in Fig. \ref{fig:Architecture}. 
Furthermore, to better align key-value pairs $(k_{i},v_{i})$ and to stabilize the training process, we also propose a novel Memory Refinement loss, of which detailed information is presented in Sec. \ref{MRlossSection}.

\begin{figure}[t]
    \centering
    \includegraphics[scale=0.5]{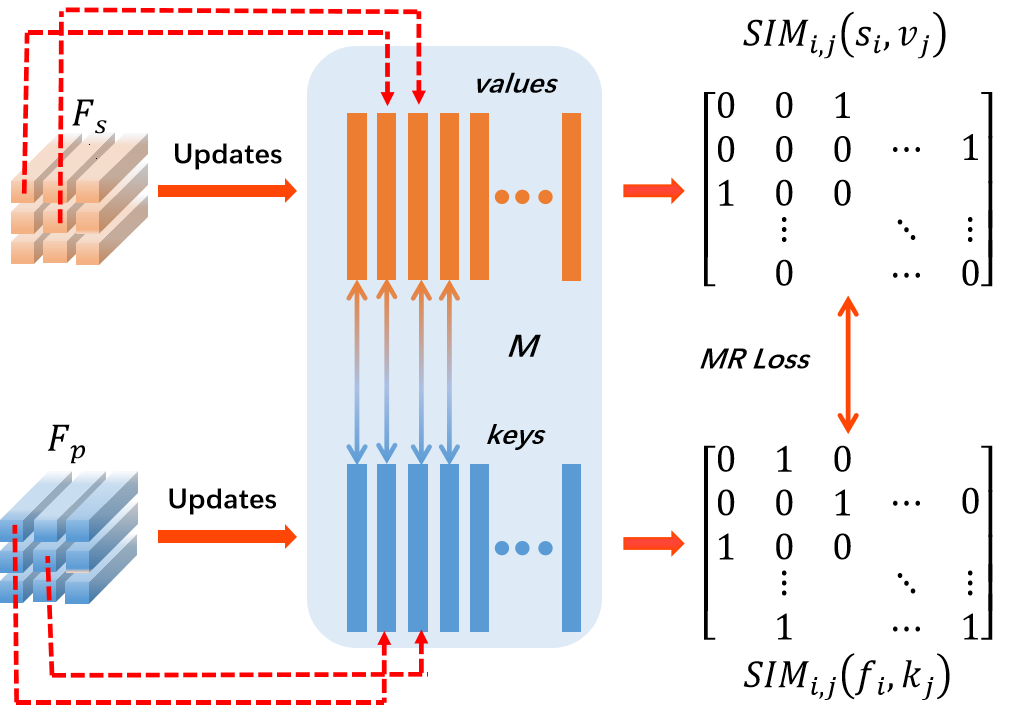}
    \caption{Illustration of the Memory Refinement Loss for feature alignment in the memory module. Two cross domain feature vectors with same spatial information tend to update one pair of key and value in particular. This refinement loss ensures that photo features and sketch features are spatially aligned.}
    \label{fig:MRLoss}
\end{figure}

After updating $M$, we retrieve synthesised sketch feature $\hat{s}_{i}$ by iterating the memory items through the query $f_{i}$. Instead of selecting the matching item $(k_{max},v_{max})$ to $f_{i}$, we use an attentive reading strategy for obtaining $\hat{s}_{i}$. We compute a weight $a_{ij}$ for each key $k_{j}$ in the $M$, and use it to calculate a weighted average for corresponding $v_{j}$ as the $\hat{s}_{i}$.

\begin{equation}
    \hat{s}_{i}=\sum_{j=1}^{K}a_{ij}\cdot v_{j},\quad a_{ij} = \frac{exp(\frac{f_{i}\cdot  k_{j}}{\left \| f_{i} \right \| \left \| k_{j} \right \| } )}{ {\textstyle \sum_{j=1}^{K}exp(\frac{f_{i}\cdot  k_{j}}{\left \| f_{i} \right \| \left \| k_{j} \right \| })} } 
\end{equation}
Finally, we aggregate all the retrieved slots to formulate final sketch representation
\begin{equation}
\hat{F}_{s}= \left \{ \hat{s}_{1}, \hat{s}_{2},\hat{s}_{3},\cdots \hat{s}_{N}\right \},\hat{s}_{i} \in \mathbb{R}^{c} 
\end{equation}
Note that during memory reading, $M$ is not updated using back-propagated gradients.

\subsection{Objective Functions}
The overall objective of the proposed model includes five loss functions: $\mathcal{L}_{ADV}$, $\mathcal{L}_{rec}$, $\mathcal{L}_{content}$, $\mathcal{L}_{style}$, and $\mathcal{L}_{MR}$. 
In this section, we present these loss functions in detail.
\subsubsection{Adversarial learning} The adversarial loss is leveraged to correctly distinguish the real sketches or generated sketches. As mentioned in Sec. \ref{networkArchitectures}, following \cite{isola2017image}, the adversarial loss is formulated as:
\begin{equation}
\begin{aligned}
    \mathcal{L}_{ADV} &= E[logD(X,Y)] \\
                    &+ E[log(1-D(X,G(X))]
\end{aligned}
\end{equation}
where X,Y denote the source photos, target sketches, G,D denote the generator and discriminator.

\subsubsection{Image Reconstruction} As an image-to-image translation task, we expect that the generated sketch sketch to be as close as possible to the ground truth.
Therefore, L1 distance is employed to represent the reconstruction loss, defined as 
\begin{equation}
    \mathcal{L}_{rec} = E[\left \| Y-G(X) \right \|_{1} ]
\end{equation}

\subsubsection{Content and Style Consistency}
In order to ensure that the synthesized sketch and the target sketch have homogeneous style while synethesized sketch and the input photo have similar content, we employ pre-trained VGG-19\cite{simonyan2014very} to extract multi-level representations. 
In terms of fake sketch $G(X)$ and target sketch $Y$, we compute the L2 distance of features from VGG-19 after pool1 and pool2 as Style Loss:
\begin{equation}
\mathcal{L}_{style} = E[ \sum_{i=1}^{2}\left \|\varphi^{i}(Y) - \varphi^{i}(G(X)) \right \|_{2}^{2}  ]
\end{equation}
where $\varphi^{i}(\cdot)$ represets the output featurrs of VGG-19 net and $i$ denotes the $i$th pool layer of VGG-19 net. 
Additionally, we compute L2 distance of $X$ and $G(x)$ from VGG-19 after pool4 as the Content Loss:
\begin{equation}
\mathcal{L}_{content} = E[ \left \|\varphi^{4}(X) - \varphi^{4}(G(X)) \right \|_{2}^{2}  ]
\end{equation} 
The style loss and content loss make training procedure more stable.

\subsubsection{Memory Refinement}
\label{MRlossSection}
With the self-supervised learning strategy, MOST-Net can realise the face sketch synthesis to a certain extent. 
However, the training process of $M$ tend to be unstable and time-consuming. 
Thus, we design a novel Memory Refinement Loss (MR Loss) for feature alignment in the memory module to help training. 
As shown in Fig. \ref{fig:MRLoss}, we define that with the same subscript $i$ , $f_{i}$ and $s_{i}$ should update one key-value pair simultaneously. 
Therefore, the most relevant key $k_{i}$ of $f_{i}$ and the most relevant value $v_{j}$ of $s_{i}$ also share same subscript $j$. Formally, the similarity permutation matrix could formulated as:
\begin{equation}\label{7}
SIM_{i,j}(f_{i},k_{j})=\begin{cases}1, k_{j} = k_{max}
 \\0,  k_{j} \ne k_{max}
\end{cases}  
\end{equation}
as well as $SIM_{i,j}(s_{i},v_{j})$. 
Then, we utilize the normalized L1 distance to represent MR Loss,
\begin{equation}
    \mathcal{L}_{MR} = E[\left \|SIM(f_{i},k_{j})-SIM(s_{i},v_{j})\right \|_{1} ]
\end{equation}

\subsubsection{Full Objective} The overall loss for training MOST-Net is 
\begin{equation}\label{16}
\begin{aligned}
    \mathcal{L}_{total}=\lambda_{1}\mathcal{L}_{ADV} &+ 
    \lambda_{2}\mathcal{L}_{rec}  + 
    \lambda_{3}\mathcal{L}_{style} \\&+
    \lambda_{4}\mathcal{L}_{content} + 
    \lambda_{5}\mathcal{L}_{MR} 
\end{aligned}
\end{equation}
where$\lambda_{1}$, $\lambda_{2}$, $\lambda_{3}$, $\lambda_{4}$, and $\lambda_{5}$ are loss weights. Besides, the generator G and the discriminator D could be optimized by:
\begin{equation}
    \underset{G}{min} \underset{D}{max}\mathcal{L}_{total}  
\end{equation}

\begin{table*}[!htbp]
\centering
\caption{Quantitative comparison results with SOTA models on CUFS and CUFSF test set. The $\uparrow$ indicates the higher is better, while $\downarrow$ indicates the lower is better. The results show that our method achieves the state-of-the-art performance.}
\begin{tabular}{|l|cc|cc|cc|cc|cc|}
\hline
\multirow{2}{*}{\begin{tabular}[c]{@{}l@{}}Methods/Years\\ Dataset\end{tabular}} &
  \multicolumn{2}{c|}{FSIM$\uparrow$} &
  \multicolumn{2}{c|}{SSIM$\uparrow$} &
  \multicolumn{2}{c|}{LPIPS (SqueezeNet)$\downarrow$ } &
  \multicolumn{2}{c|}{LPIPS (AlexNet)$\downarrow$ } &
  \multicolumn{2}{c|}{LPIPS (VGG-16)$\downarrow$ } \\ \cline{2-11} 
                          & CUFS            & CUFSF           & CUFS   & CUFSF  & CUFS   & CUFSF           & CUFS   & CUFSF           & CUFS   & CUFSF  \\ \hline
CycleGAN\cite{zhu2017unpaired} (2017)            & 0.6829          & 0.7011 & 0.4638 & 0.3753 & 0.1863 & 0.1617 & 0.2776 & 0.2234 & 0.3815 & 0.3787 \\ \hline
Pix2Pix\cite{isola2017image} (2017)            & 0.7356          & 0.7284 & 0.4983 & 0.4204 & 0.1262 & 0.1422 & 0.1876 &\textbf{0.1932} & 0.3217 & 0.3551 \\ \hline
Col-cGan\cite{zhu2019deep} (2019)                 &-        & - & 0.5244     & 0.4224      & - & - & - & - & - & - \\ \hline
KT\cite{zhu2019face} (2019)                 & 0.7373          & \textbf{0.7311} & -      & -      & 0.1688 & 0.1740 & 0.2297 & 0.2522 & 0.3483 & 0.3743 \\ \hline
KD+\cite{zhu2020knowledge} (2020)                & 0.7350          & 0.7171 & -      & -      & 0.1471 & 0.1619 & 0.1971 & 0.2368 & 0.3052 & 0.3550 \\ \hline
SCAGAN\cite{yu2020toward} (2020)             & 0.7086          & 0.7270 & -      & -      & 0.1722 & 0.1500 & 0.2408 & 0.2188 & 0.3627 & 0.3536 \\ \hline

Sketch-Trans\cite{zhu2021sketch} (2021) & 0.7350          & 0.7259 & -      & -      & 0.1233 & \textbf{0.1349} &\textbf {0.1807} & 0.1971 & 0.3019 & 0.3400 \\ \hline
\textbf{MOST-Net (ours)} &
  \textbf{0.7394} &
  0.7167 &
  \textbf{0.5433} &
  \textbf{0.4363} &
  \textbf{0.1207} &
  0.1386 &
  0.1837 &
  0.2013 &
  \textbf{0.3006} &
  \textbf{0.3374} \\ \hline
\end{tabular}
\label{tab:resultsl}
\end{table*}

\section{Experiments}
\subsection{Implements Details}
Our network is trained from scratch without any prior knowledge through an end-to-end training strategy. The Adam optimizer is adopted with $\beta_{1}=0.9$ and $\beta_{2}=0.999$. The learning rates are set to $0.001$ and $0.0004$ for the discriminator and generator respectively. For the hyper-parameters in Eq. \ref{16}, we set $\lambda_{1}$, $\lambda_{2}$, $\lambda_{3}$, $\lambda_{4}$, and $\lambda_{5}$ to 1, 200, 40, 40, and 10. Additionally, the decay rate in Eq. \ref{6} is set to 0.999. Since we need a small decay rate for stabilizing the memory updating procedure, it takes more than 1000 epochs with a batch size of 16 for our network to converge. 
\subsection{Datasets and Evaluation Metrics}
We conduct experinments on the CUFS dataset\cite{tang2003face} and the CUFSF dataset\cite{zhang2011coupled}.
In the CUFS dataset, there are 606 faces, of which 188 faces from the CUHK student database, 123 faces from the AR database, and 295 faces from the XM2VTS database. 
For each paired sample, the face photo is taken in natural lightning conditions while the sketch is drawn by the artist. 
The CUFSF dataset contains 1194 face-sketches pairs. However, all photos are obtained under illumination variations as illustrated in Fig. \ref{fig:task}. 
For both datasets, the whole images are cropped to $200\times 250$. Thus, widely used reshaping and padding strategies \cite{yu2020toward} for face sketch synthesis are adopted to expand the input image size to $256\times256$.

The performance of sketch synthesis is measured by multiple metrics. 
We first employ the Feature Similarity Index Metric (FSIM)\cite{zhang2011fsim} and Structural Similarity Index Metric (SSIM)\cite{wang2004image} to evaluate the quality of synthesized sketches. 
Furthermore, for assessing the similarity between the ground-truth and synthesized sketch from human perspectives, we choose the Learned Perceptual Image Patch Similarity (LPIPS)\cite{zhang2018unreasonable}. 
In this paper, three different perceptual similarity models are used as the baseline of LPIPS, including SqueezeNet\cite{iandola2016squeezenet}, AlexNet\cite{krizhevsky2012imagenet}, and VGGNet\cite{simonyan2014very}.



\begin{figure}[!t]
\centering
\includegraphics[scale=0.48]{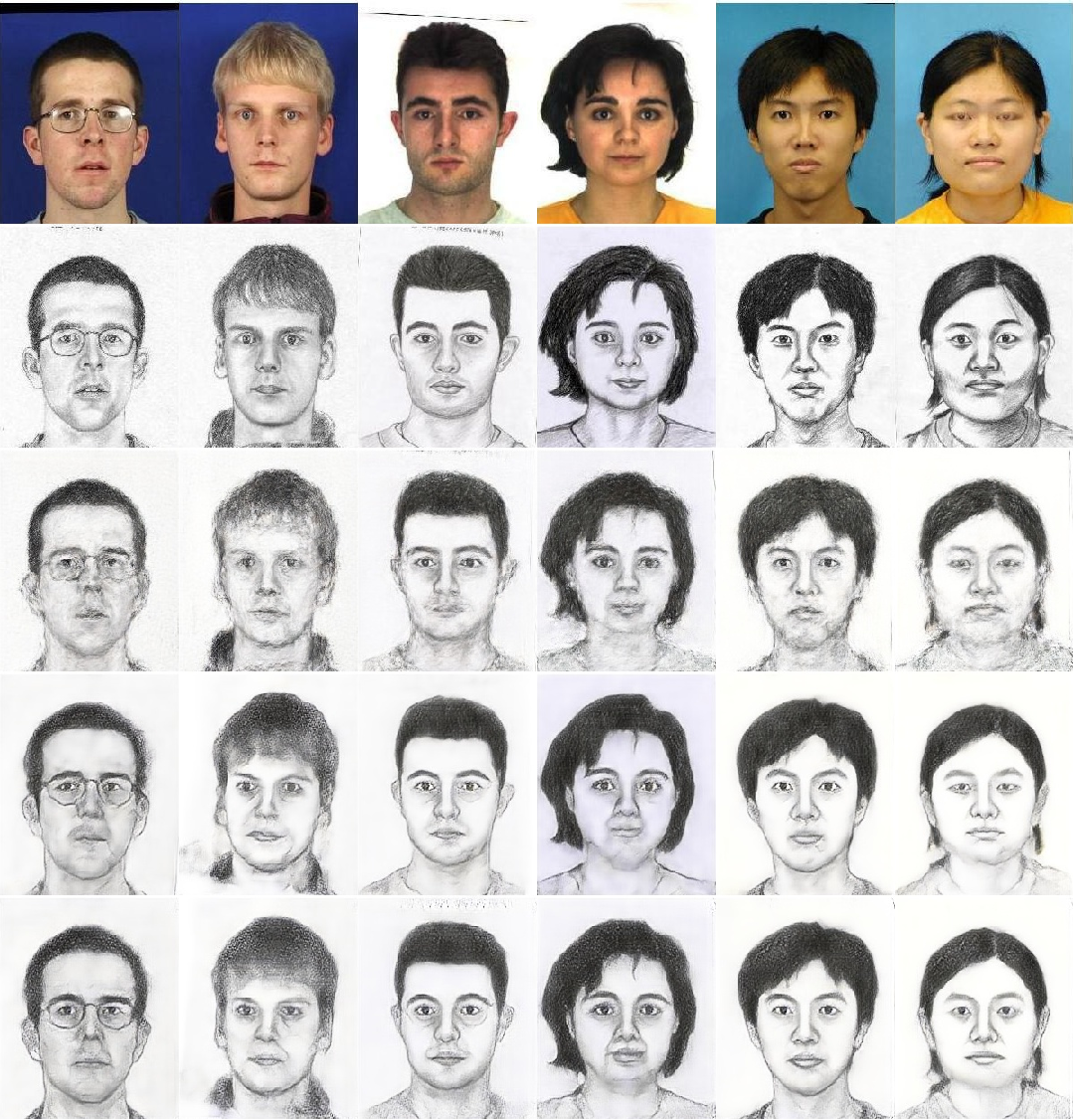}
\caption{Ablation studies of synthesized sketches on the CUFS dataset. For each sample, the first row is the origin face, while the second row is the targeted sketch, followed by sketches generated by Pix2Pix, MOST-Net Without MR Loss, and MOST-Net with MR Loss respectively.}
\label{fig:ablation}
\end{figure}

\subsection{Experiment Results and Analysis}

The quantitative results of MOST-Net on the CUFS and the CUFSF test sets are reported in Table. \ref{tab:resultsl}. The proposed method achieves the best performance on the metrics of SSIM, which suggests that the sketch generated by our method holds the highest structural similarity with the ground-truth sketch. Because we spatially align the cross-domain features, it is reasonable for sketches to have similar structures. On the CUFS dataset, we obtain the state-of-the-art performance among all face sketch synthesis methods on the indicators of FSIM, LPIPS-SqueezeNet, and LPIPS-VGG16. Additionally, Our method 
slightly increases the previous best LPIPS-AlexNet from 0.1807 to 0.1837.
On the CUFSF dataset, there is not any dominant method for all perspectives. 
We analyze the reason is that samples of the CUFSF dataset are non-aligned and with low-fidelity. 
The performance of the network targeting image-to-image translation on this data set is not stable and consistent. 
However, our method achieves the optimal result of LPIPS-VGG16 and sub-optimal results of LPIPS-SqueeeNet and LPIPS-AlexNet. However, our method decreases the previous best FSIM from 0.7311 to 0.7167 on the CUFSF dataset. 
Because LPIPS assesses similarity between two images from the human perspective \cite{zhang2018unreasonable}, we consider LPIPS as a more meaningful evaluation metric for face sketch synthesis task. 
Therefore, we can conclude that our method could integrally construct sketches from face photos with high-fidelity and rich details and achieves state-of-the-art performances in this task.

\begin{table}[ht]
\centering
\caption{Results of ablation study on CUFS dataset. The $\uparrow$ indicates the higher is better, while $\downarrow$ indicates the lower is better.}
\label{ablations}
\begin{tabular}{|l|l|l|l|}
\hline
Metrics & Baseline & \begin{tabular}[c]{@{}l@{}}MOST-Net \\ w/o MR Loss\end{tabular} & \begin{tabular}[c]{@{}l@{}}MOST-Net \\ w/ MR Loss\end{tabular} \\ \hline
FSIM $\uparrow$                                                         & 0.7356 & 0.735 & \textbf{0.7394} \\ \hline
SSIM $\uparrow$                                                         & 0.4983 & 0.5431 & \textbf{0.5433} \\ \hline
\begin{tabular}[c]{@{}l@{}}LPIPS $\downarrow$  \\ (SqueezeNet)\end{tabular} & 0.1362 & 0.1303 & \textbf{0.1207} \\ \hline
\begin{tabular}[c]{@{}l@{}}LPIPS $\downarrow$ \\ (AlexNet)\end{tabular}    & 0.1876 & 0.1882 & \textbf{0.1837} \\ \hline
\begin{tabular}[c]{@{}l@{}}LPIPS $\downarrow$ \\ (VGG-16)\end{tabular}     & 0.3217 & 0.312 & \textbf{0.3006} \\ \hline
\end{tabular}
\end{table}

\subsection{Ablation Study}
Furthermore, we conduct an ablation study to verify the effectiveness of the proposed memory refinement loss. Because our method utilizes a similar backbone of U-Net \cite{ronneberger2015u} as the Pix2Pix network, we consider Pix2Pix as the baseline. Then, we remove MR Loss from the MOST-Net to investigate its value. From Fig. \ref{fig:ablation} we can observe that sketches generated by our method are more vivid than Pix2Pix. Moreover, as illustrated in Tab. \ref{ablations}, MOST-Net without MR Loss drastically increases SSIM from 0.4983 to 0.5431, which proves the capability of the network itself. Additionally, MR Loss slightly optimizes all indicators and further improves the sketch synthesis performance.

\subsection{Limitation and Future Work}
Although our method obtains the state-of-the-art performance for the face sketch synthesis task, there is still a lot of value left to be extracted. The primary limitation of this study is that aligned training images are needed. As we can observe in Tab. \ref{tab:resultsl}, our method is overwhelmed by the CUFSF dataset. It is natural because we update key-value pairs through spatially aligned feature vectors. An alternative way to overcome this drawback may refer to a semi-supervised or weak-supervised learning strategy, which we are going to further study.

\section{Conclusion}
In this paper, we propose a Memory Oriented Style Transfer Network (MOST-Net) for face sketch synthesis. 
Specially, we design a novel end-to-end domain transfer framework based on an external memory module.
Meanwhile, we present a Memory Refinement loss for feature alignment in the memory dictionary and stabilizing the training procedure. 
We conduct extensive experiments on the CUFS and CUFSF datasets. 
The results show that our proposed MOST-Net achieves state-of-the-art performance. 
Additionally, our method could be adopted as a backbone network in other image domain transfer tasks. 
In the future, we will conduct more qualitative and quantitative experiments to verify the feasibility of our model for general purposes. 
A more complete version of this research will be realized in the future.

\section{Acknowledgement}
This work is founded by the National Natural Science Foundation of China under Grant U1836217.

\newpage

\bibliographystyle{IEEEtran}
\bibliography{IEEEfull,references}


\end{document}